\documentclass[11pt,a4paper]{article}
\usepackage[hyperref]{acl2021}
\usepackage{times}

\usepackage{latexsym}
\usepackage{graphicx}
\usepackage{booktabs}
\usepackage{multirow}
\usepackage{amsmath}
\usepackage{amssymb}
\usepackage{setspace}
\usepackage{bm}

\usepackage{url}

\usepackage{microtype}

\aclfinalcopy 

\title{SemEval-2021 Task 4: Reading Comprehension of Abstract Meaning}

\newcommand*{\affaddr}[1]{#1} 
\newcommand*{\affmark}[1][*]{\textsuperscript{#1}}
\newcommand*{\email}[1]{\texttt{#1}}

\author{%
 Boyuan Zheng\affmark[2]\thanks{\quad This work was performed when Boyuan Zheng visited Queen's University.} , Xiaoyu Yang\affmark[1], Yu-Ping Ruan\affmark[3],  Zhenhua Ling\affmark[3], 
 
 \\ \textbf{Quan Liu\affmark[3]}, \textbf{Si Wei\affmark[4]}, \textbf{Xiaodan Zhu\affmark[1]}\\
\affaddr{\affmark[1]Queen's University, Kingston, Canada;}
\affaddr{\affmark[2]Northeastern University, Shenyang, China}\\
\affaddr{\affmark[3]University of Science and Technology of China;}
\affaddr{\affmark[4]iFlytek Research, Hefei, China}\\
\email{steven.zheng010@gmail.com};
\email{xiaoyu.yang@queensu.ca}; \\
\email{\{quanliu,zhling\}@ustc.edu.cn};
\email{siwei@iFlytek.com.cn};\\
\email{xiaodan.zhu@queensu.ca}
}

\date{}

\begin{document}
\maketitle
\begin{abstract}
This paper introduces the SemEval-2021 shared task 4: Reading Comprehension of Abstract Meaning (ReCAM). This shared task is designed to help evaluate the ability of machines in representing and understanding abstract concepts.
Given a passage and the corresponding question, a participating system is expected to choose the correct answer from five candidates of abstract concepts in a cloze-style machine reading comprehension setup.
Based on two typical definitions of \textit{abstractness}, i.e., the \textit{imperceptibility} and \textit{nonspecificity},
our task provides three subtasks to evaluate the participating models. Specifically, Subtask 1 aims to evaluate how well a system can model concepts that cannot be directly perceived in the physical world.
Subtask 2 focuses on models' ability in comprehending \textit{nonspecific} concepts located high in a hypernym hierarchy given the context of a passage. 
Subtask 3 aims to provide some insights into models' generalizability over the two types of abstractness. 
During the SemEval-2021 official evaluation period, we received 23 submissions to Subtask 1 and 28 to Subtask 2. The participating teams additionally made 29 submissions to \mbox{Subtask 3}. The leaderboard and competition website can be found at \url{https://competitions.codalab.org/competitions/26153}. The data and baseline code are available at \url{https://github.com/boyuanzheng010/SemEval2021-Reading-Comprehension-of-Abstract-Meaning}.
\end{abstract}

\section{Introduction}

Humans use words with abstract meaning in their daily life. In the past, research efforts have been exerted to better understand and model abstract meaning ~\cite{turney2011literal,theijssen2011difficulty,mark2008hierarchies,spreen1966abstract}. Modelling abstract meaning is closely related to many other NLP tasks such as reading comprehension, metaphor modelling, sentiment analysis, summarization, and word sense disambiguation.

In the past decade, significant advancement has been seen in developing computational models for semantics, based on deep neural networks. In this shared task, we aim to help assess the capability of the state-of-the-art deep learning models on representing and modelling abstract concepts in a specific reading comprehension setup.


We introduce SemEval-2021 Task 4, Reading Comprehension of Abstract Meaning (\textbf{ReCAM}). Specifically, we design this shared task by following the machine reading comprehension framework ~\cite{karl2015teach, onishi2016did, hill2015cbtest}, in which computers are given a passage ${D_i}$ as well as a human summary ${S_i}$ to comprehend. If a model can digest the passage as humans do, we expect it to predict the abstract word used in the summary, if the abstract word is masked. 
Unlike the previous work that requires computers to predict concrete concepts, e.g., named entities, in our task we ask models to fill in abstract words removed from human summaries. During the SemEval-2021 official evaluation period, we received 23 submissions to Subtask 1 and 28 submissions to Subtask 2. The participating teams additionally made 29 submissions to Subtask 3. In this paper, we induce the shared task and provide a summary for the evaluation.

\section{Task Description}
\label{sec:overview}
\vspace{-2mm}
We organize our shared task based on two typical definitions of \textit{abstractness}, named as \textit{imperceptibility} and \textit{nonspecificity} in this paper, implemented in Subtask 1 and Subtask 2, respectively. \mbox{Subtask 3} further evaluates models' generalizability over the two definitions of abstractness.

\vspace{-2mm}

\subsection{Subtask 1: ReCAM-Imperceptibility}
In one definition~\cite{turney2011literal, theijssen2011difficulty, spreen1966abstract}, concrete words refer to things, events, and properties that humans can directly perceive with their senses, e.g., \textit{trees} and \textit{flowers}. In contrast, abstract words refer to ``ideas and concepts that are distant from immediate perception", e.g.,~\textit{objective}, \textit{culture}, and \textit{economy}. In Subtask 1, we perform reading comprehension on \textit{imperceptible} abstract concepts, named as \textbf{ReCAM-ImPerceptibility}. Table~\ref{tab:impercept-example} shows an example. 






\begin{table}[!t]
\setlength{\belowcaptionskip}{-0.3cm}
\centering
\begin{tabular}{|p{1.1cm}|p{6cm}|}
  \hline
  {\footnotesize{
  \bf{Passage}
  }} & 
  {\footnotesize{
    ... Observers have even named it after him, ``Abenomics". It is based on three key pillars of monetary policy to ensure long-term sustainable growth in the world's third-largest economy, with fiscal stimulus and structural reforms. In this weekend's upper house elections, ....
  }}\\
  \hline
  {\footnotesize{\bf{Question}}} &  {\footnotesize{Abenomics: The \textit{\textbf{@placeholder}} and the risk.}}\\
  \hline
  {\footnotesize{\bf{Answer}}} &\footnotesize{
  (A) chance \quad   (B) prospective\quad(C) government \quad\textbf{(D) objective}\quad 
   (E) threat}\\
  \hline
\end{tabular}
\caption{An example for Subtask 1. The correct answer to the question is \textbf{\textit{objective}}.}\label{tab:impercept-example}
\end{table}

\subsection{Subtask 2: ReCAM-NonSpecificity}
The second typical definition of abstractness is based on~\textit{nonspecific} concepts~\cite{ theijssen2011difficulty,spreen1966abstract}. Compared to specific concepts such as~\textit{groundhog} and~\textit{whale}, words such as~\textit{vertebrate} are regarded as more \textit{abstract}. Our Subtask 2, named as \textbf{ReCAM-NonSpecificity}, is designed based on this viewpoint. We will discuss how the datasets are constructed in Section~\ref{sec:data}.





\subsection{Subtask 3: ReCAM-Cross}
In this subtask, participants are asked to submit their predictions on the test data of Subtask 2, using models trained on the training data of Subtask 1, and vice versa. This subtask aims to demonstrate models' generalizability between modelling the two typical definitions of abstractness.

\section{Data Construction}
\label{sec:data}
We develop our multi-choice machine reading comprehension datasets based on the XSum summarization dataset~\cite{shashi2018xsum}. We first locate words with abstract meaning using our abstractness scorers. Then we perform data filtering to select our target words to construct our datasets.



\subsection{The XSum Data}
By collecting online articles from the British Broadcasting Corporation (BBC), ~\citet{shashi2018xsum} developed a large-scale text summarization dataset, XSum, in which each article has a single sentence summary. We developed our ReCAM dataset based on XSum. 


\subsection{Finding \textit{Imperceptible} Concepts}
\paragraph{Abstractness Scorer for Imperceptibility}
\label{sec:appendix-mrc}
Following~\newcite{turney2011literal}, we use the MRC Psycholinguistic Database \cite{coltheart1981mrc}, which includes 4,295 words rated with a degree of abstractness by human subjects, to train our abstractness scorer for \textit{imperceptibility}. The rating of the words in the MRC Psycholinguistic Database ranges from 158 (highly abstract) to 670 (highly concrete). 
We linearly scale the rating to the range of 0 (highly abstract) to 1 (highly concrete). The neural regression model accepts fixed Glove embedding \cite{pennington2014glove} as input and predicts the abstractness rating score between 0 and 1. Our regression model is a three-layer network that consists of two non-linear hidden layers with the ReLU activation and a sigmoid output layer. The mean square error (MSE) is used as the training loss.

To test the regression model's performance, we randomly split the MRC Psycholinguistic Database into train and test set with the size of 2,148 and 1,877, respectively. 
Table \ref{tab:mrc_regression} shows the final performance of the neural regression model on the MRC database. We use the Pearson correlation between ratings predicted by models and original ratings from MRC as the evaluation metric. We can see that the regression model achieves high correlation coefficients (the higher, the better), i.e., 0.934 and 0.835, on the training and test set. The correlations are significant ($p$-values are smaller than $10^{-5}$), reflecting the quality of our models in finding abstract words. Note that \newcite{turney2011literal} report a correlation score of $0.81$ on their MRC test set. Their training-test split is unavailable, so we run cross-validation here in our experiment. 
The scorer can then be used to assign an \textit{imperceptibility} score to a word that is not in the MRC Psycholinguistic Database.

\begin{table}[!t]
\centering
\begin{tabular}{|l|ccc|}
  \hline
   & \#samples & Pearson $r$ & $p$-value \\
   \hline
  train & 2,148 & 0.934 & $p<10^{-5}$ \\
  test & 1,877 & 0.854 & $p<10^{-5}$ \\
  \hline
\end{tabular}
\caption{Fitting performance of neural regression model on the MRC database.}\label{tab:mrc_regression}
\end{table}

Using the abstractness scorer described above, we assign an abstractness value to each word in summaries and select words with a value lower than 0.35 as the candidates for our \textit{target words} (words that will be removed from the summaries to construct questions). We only consider content words as potential target words, i.e., nouns, verbs, adjectives, and adverbs. For this purpose, we use part-of-speech tagging model~\cite{qi-etal-2018-universal} implemented in Stanza \cite{qi2020stanza}.


\subsection{Finding \textit{Nonspecific} Concepts}
\paragraph{Nonspecificity Scorer}
Following the work of~\citet{mark2008hierarchies}, we assign a \textit{nonspecificity} score to a word token based on the hypernym hierarchy of WordNet~\cite{miller1998wordnet}. 
Specifically, the root of the hierarchy is at level 0 and regarded as the most abstract. The abstractness of a node in the hierarchy is measured by the maximal length of its path to the root. The hypernym level in WordNet is between 0 and 17. For each word token in summaries, we use Adapted Lesk Algorithm \cite{satan2002alesk} to label the sense since the WordNet hypernym hierarchy works at the sense level. Since a summary sentence may be short, we concatenate each summary sentence with the corresponding passage for word sense disambiguation. Built on this, each token, which is labelled with a sense, receives an abstractness score based on the WordNet hierarchy. 

Using the \textit{nonspecificity} scorer, we assign an \textit{nonspecificity} value to each word in summaries and select words with a value smaller than six as the candidate target words. The targets words will be nouns and verbs since the hypernym hierarchy in WordNet~\cite{marciniak-2020-wordnet} consists of these two POS types.

\subsection{Filtering}
We aim to avoid developing simple questions. For example, if a target word also appears in the passage, it is likely that a model can easily find the answer without the need to understand the passage in depth.  



\paragraph{Filtering by Lemmas}

We lemmatized passages and summaries. If a lemma appears both in a summary and the corresponding passage, the lexemes of the lemma will not be considered as target words. Note that a strict filter may exclude some good candidates for target words but helps avoid introducing many simple questions. 



\paragraph{Filtering by Synonyms and Antonyms}
\label{synonym_antonym}

For a word in a summary, if a synonym or antonym of the word appears in the corresponding passage, we will not consider this word to be our target word.  We use WordNet~\cite{marciniak-2020-wordnet} to derive synonyms and antonyms. Instead of using word sense disambiguation (WSD), for a word $w_i$ in a summary, we use all senses of this word and add all synonyms and antonyms into a pool. Only if none of the words in the pool appear in the passage, we consider $w_i$ as a candidate target word. Otherwise, we will not use $w_i$ to construct a question for this passage-summary pair.


\paragraph{Filtering by Similarity}
\label{context_similarity}
We further filter words by similarity. For each candidate target word in a summary and each word in the passage, we calculate similarity and use that to perform further filtering. 



We use 300-dimension GloVe word embedding trained on 840 billion tokens~\cite{pennington2014glove}. We calculate the cosine similarity between a candidate target word and a passage word. For contextual embedding, we embed each sentence in a passage as well as the summary into a context-aware representation matrix using the BERT-large uncased language model. Then, we calculate the similarity between each passage token and question token with the cosine similarity. If the similarity is higher than 0.85, we will not consider the involved summary words as candidate target words.


\subsection{Constructing Multiple Choices}
We train machine reading comprehension models using the data built so far to generate four choices for each question. Together with the ground-truth (the target word identified above and removed from the human summary), we have five choices/options for each question. In our work, we propose to use three models, Gated-Attention Reader~\cite{karl2015teach}, Attentive Model and Attention Model with Word Gloss to generate the candidate options. Please find details of the models in  Appendix~\ref{gated_attention_reader} and  Appendix~\ref{attentive_model} as well as the training details in  Appendix~\ref{training_details}.

We adopt the idea of k-fold cross validation to train the above mentioned three models to generate candidate answer words. Specifically, we split the data into 4 folds. Each time, we train the baseline models on 3 folds of data and use the trained models to predict candidate words on the remaining 1-fold data. With 4-fold
iteration, we obtain predication of each model on the entire data. The performance of the three baseline models are listed in Table~\ref{tab:results_1} for Subtask 1 and Table~\ref{tab:results_2} for Subtask 2, using several typical retrieval-based evaluation metrics.

For each target word that has been removed from the corresponding summary sentence (again, a question is a summary sentence containing a removed target word), we collect top-10 words predicted by each of the three models. In this way, we can collect a candidate word pool of 30 predicted word tokens for each removed target word. To avoid including multiple correct choices for each question, we adopt synonym and context similarity filtering methods described in Section~\ref{synonym_antonym}. Specifically we first calculate similarity between the ground-truth target word and each word type in the pool. We exclude a word type from the multiple choices if its similarity to the ground-truth is higher than 0.85. In addition, we also exclude synonyms of the ground-truth target word. For the remaining word tokens in the pool, we select four most frequent word types (a word type may have multiple tokens in the pool). Together with the ground-truth word, we obtain five choices for each question.

\begin{table}[!t]
\centering
\begin{tabular}{|p{1.56cm}|p{1cm}p{1cm}p{1cm}p{1cm}|}
  \hline
   & MRR & R@1 & R@5 & R@10 \\
   \hline
  GAReader & 0.245 & 0.175& 0.314 & 0.378 \\
  \hline
  AttReader & 0.235 & 0.167 & 0.300 & 0.363\\
  \ \ \ +gloss &0.179 &0.123 &0.227 &0.276\\
  \hline 
  
\end{tabular}
\caption{Three baseline models are used to generate candidate multiple choices for Subtask 1. The table shows their performance on the XSum dataset, evaluated with MRR\cite{craswell2009mean}, Recall@1, Recall@5, and Recall@10.}\label{tab:results_1}
\end{table}

\begin{table}[!t]
\centering
\begin{tabular}{|p{1.56cm}|p{1cm}p{1cm}p{1cm}p{1cm}|}
  \hline
   & MRR & R@1 & R@5 & R@10 \\
   \hline
  GAReader &0.343  &0.268  &0.422  &0.484 \\
  \hline
  AttReader &0.348  &0.273  &0.424  &0.490 \\
  \ \ \ +gloss &0.228 &0.166 &0.286 &0.345\\
  \hline

\end{tabular}
\caption{Three baseline models are used to generate candidate multiple choices for Subtask 2. The table shows their performance on the XSum dataset, evaluated with MRR, Recall@1, Recall@5, and Recall@10.}\label{tab:results_2}
\end{table}

\subsection{Further Quality Control}
We further make the following efforts to remove noise in the dataset and improve the datasets' quality. We observe that up to now, there are mainly two kinds of noise in our dataset: 1) some target words cannot be inferred solely based on the corresponding passage; 2) more than one of the multiple choices are correct answers. 

The first issue is  mainly related to the property of the XSum dataset, in which the first sentence of a passage is used as the summary. The second type of problems are often caused by our automatic generation method. Although we have applied strict rules in Section~\ref{context_similarity} to handle this, among a small portion of the resulting data, multiple potentially correct answers still exist in candidate answers.

To further ensure the quality of our dataset, we invite workers in Amazon Mechanical Turk to perform further data selection. Each annotator needs to follow the procedure of Appendix~\ref{annotation_script} to answer the question and annotate relevant information, with which further data selection is applied. To ensure quality, we only include workers from English-speaking countries and only if their previous HITs' approval rates are above 90\%. To see more details about this process, please refer to  Appendix~\ref{annotation_selection}.


\subsection{ReCAM Data Statistics}
\label{sec:dataset-size}
Table \ref{tab:data-size-table} lists the size of our ReCAM datasets, i.e., numbers of questions. For example, in total Subtask 2 has 6,186 questions, which are split into training/development/test subsets.

\begin{table}[!h]
\centering
\begin{tabular}{|l|ccc|}
  \hline
  Dataset& Subtask 1 &Subtask 2 & Total\\
  \hline
  Train&3,227 &3,318 &6,545\\
  Dev&837 &851 &1,688\\
  Test&2,025 &2,017 &4,042 \\
  \hline
  Total&6,089 &6,186 & 12,275\\
  \hline
\end{tabular}
\vspace*{-2mm}
\caption{Size of the ReCAM Dataset.}\label{tab:data-size-table}
\end{table}

\section{Systems and Results}
Our shared task received 23 submissions to Subtask 1, 28 submissions to Subtask 2, and 29 submissions to Subtask 3. We use \textit{accuracy} as the evaluation metric for the three subtasks.

In general, most participating teams use pre-trained language models in their systems such as BERT~\cite{bert}, ALBERT~\cite{albert}, DistilBERT~\cite{distilbert}, RoBERTa~\cite{roberta}, ELECTRA~\cite{electra}, DeBERTa~\cite{deberta}, XLNet~\cite{xlnet}, T5~\cite{t5}.
 Data augmentation, external knowledge resources, and/or transfer learning are additionally used by many teams to further enhance their model performance.

\subsection{Subtask 1: ReCAM-Imperceptibility}
Table ~\ref{tab:subtask_1_results} shows all the official submissions and most of them outperform the baseline model. The baseline used for Subtask 1 is the Gated-Attention (GA) Reader~\cite{dhingra2016gated}. The GA Reader uses a multi-layer iterated architecture with a gated attention mechanism to derive better query-aware passage representation. The motivation behind using GA Reader is to have a simple comparison between our task and the CNN/Daily Mail reading comprehension dataset since GA Reader achieves reasonably good  performance on the CNN/Daily Mail reading comprehension dataset. 

Note that the last column of the table lists the accuracy (\textit{Acc. Cross}) for models trained on the \mbox{Subtask 2} training data and tested on the Subtask 1 testset. We will discuss those results later in Section~\ref{sec:subtask3}.

\begin{table}[!t]
\renewcommand\arraystretch{1.25}
\centering
\begin{tabular}{|l|p{2.5cm}|l|c|}
  \hline
  Rank & Team  & Acc & Acc. Cross\\
  \hline
  -&GA Reader&25.1& -\\
  \hline
  1& SRC-B-roc &95.1   & 91.8 ($\downarrow$ 3.3)\\
  2&PINGAN-Omini-Sinitic &93.0 &91.7 ($\downarrow$ 1.3)\\ 
  3&ECNU-ICA-1 &90.5 &88.6($\downarrow$ 1.9)\\
  4&tt123 &90.0 &86.2($\downarrow$ 3.8)\\
  5&cxn &88.7 & -\\
  6&nxc &88.6 & 74.2($\downarrow$ 14.4)\\
  7& ZJUKLAB &87.9  & -\\
  8&IIE-NLP-Eyas &87.5 &82.1($\downarrow$ 5.4)\\
  9&hzxx1997 &86.7 & -\\
  10&XRJL &86.7 &81.8($\downarrow$ 4.9)\\
  11&noobs &86.2 &78.6($\downarrow$ 7.6)\\
  12&godrevl &83.1 & -\\
  13& ReCAM@IITK &  82.1 &80.7($\downarrow$ 1.4)\\
  14&DeepBlueAI &81.8 &76.3($\downarrow$ 5.5)\\
  15&LRG &75.3 &61.8($\downarrow$ 13.5)\\
  16&xuliang &74.7 & -\\
  17&Llf1206571288 &72.8 & -\\
  18&Qing &71.4 & -\\
  19& NEUer &  56.6 &51.8($\downarrow$ 4.8)\\
  20&CCLAB &46.3 &35.2($\downarrow$ 11.1)\\
  21&UoR &42.0 & 39.4($\downarrow$ 2.6)\\
  22&munia &19.3 & -\\
  23&BaoShanCollege &19.0 & -\\
  \hline
\end{tabular}
\vspace*{-2mm}
\caption{Official results of Subtask 1 and Subtask 3. \textit{Acc} is the accuracy of the models trained on the \mbox{Subtask 1} training data and tested on the Subtask 1 testset. \textit{Acc. cross} is the accuracy of models trained on the \mbox{Subtask 2} training data and tested on the Subtask 1 testset.}\label{tab:subtask_1_results}
\end{table}

The best result in Subtask 1 was achieved by team \texttt{SRC-B-roc}~\cite{sansung} with an accuracy of 0.951. The system was built on a pre-trained ELECTRA discriminator and it further applied upper attention and auto-denoising mechanism to process long sequences. The second-placed system, \texttt{PINGAN omini-Sinitic}~\cite{pingan}, adopted an ensemble of ELECTRA-based models with task-adaptive pre-training and a mutli-head attention based multiple-choice classifier. \texttt{ECNU-ICA-1}~\cite{ecnu} ranked third in this subtask with a knowledge-enhanced Graph Attention Network and a semantic space transformation strategy.

Most teams in Subtask 1 utilize pre-trained language models (PLM), like BERT~\cite{bert}, ALBERT~\cite{albert}, DistilBERT~\cite{distilbert}, RoBERTa~\cite{roberta}, ELECTRA~\cite{electra}, DeBERTa~\cite{deberta}, XLNet~\cite{xlnet}, T5~\cite{t5}.
\texttt{SRC-B-roc}~\cite{sansung} conducted an ablation study regarding the performance discrepancy of different transformers-based pre-training models. They tested BERT, ALBERT, and ELECTRA by directly fine-tuning the pre-trained LMs on the ReCAM data. ELECTRA outperforms BERT and ALBERT by large margins, which may be due to the different learning objectives of these pre-trained models.



Most participating systems performed intermediate task pre-training~\cite{pruksachatkun2020intermediate} for their language models. For example, CNN/Daily Mail dataset was selected by \texttt{ZJUKLAB}~\cite{zju} to further pre-train their language models. The CNN/Daily Mail dataset and Newsroom dataset  boost model performance on both Subtask 1 and \mbox{Subtask 2}. Data augmentation methods are also popular among participants. \texttt{ZJUKLAB}~\cite{zju} performed negative data augmentation with a language model to leverage misleading words. \texttt{IIE-NLP-Eyas}~\cite{iie} adopted template-based input reconstruction methods to augment their dataset and further fine-tuned their language models based on the dataset.

Most teams also used an ensemble of multiple pre-trained language models to further enhance model performance. \texttt{SRC-B-roc}~\cite{sansung} applied Wrong Answer Ensemble~\cite{kim2020learning} by training the model to learn the correct and wrong answer separately and ensembled them to obtain the final predictions. Stochastic Weight Averaging~\cite{izmailov2018averaging} was also performed across multiple checkpoints in the same run to achieve better generalization.

In addition, some interesting approaches were additionally used to tackle the task from different perspectives. \texttt{PINGAN omini-Sinitic}~\cite{pingan} turned the original multi-choice task into a masked-sentence classification task by adding each option to the placeholder. Noise detection methods and auto denoising methods were further proposed by adding a noise-tolerant loss. 
\texttt{ZJUKLAB}~\cite{zju} used label smoothing to encourage the activations of the penultimate layer. \texttt{ECNU-ICA-1}~\cite{ecnu} utilized a semantic space transformation strategy to convert ordinary semantic representations into abstract representations for classification.

\begin{table}[!t]
\renewcommand\arraystretch{1.25}
\begin{tabular}{|l|p{2.5cm}|l|c|}
  \hline
  Rank & Team  & Acc. & Acc. Cross\\
  \hline
  -&GA Reader&24.3 & -\\
  \hline
  1&PINGAN-Omini-Sinitic  &95.3 &94.2 ($\downarrow$ 1.1)\\
  2&SRC-B-roc &94.9 &93.9($\downarrow$ 1.0)\\ 
  3&tt123 &93.4 &85.8($\downarrow$ 7.6)\\
  4&ECNU-ICA-1 &93.0 &92.8($\downarrow$ 0.2)\\
  5&cxn &92.9 & -\\
  6&ZJUKLAB &92.8 & -\\
  7&nxc &92.7 &-\\
  8&hzxx1997 &90.2 & -\\
  9&XRJL &90.0 &87.6($\downarrow$ 2.4)\\
  10&IIE-NLP-Eyas &89.6 &84.1($\downarrow$ 5.5)\\
  11&ReCAM@IITK &87.6 &85.2($\downarrow$ 2.4)\\
  12&noobs &87.1 &82.4($\downarrow$ 4.7)\\
  13&DeepBlueAI &86.2 &80.7($\downarrow$ 5.5)\\
  14&xuliang &81.0 & -\\
  15&LRG &77.8 &65.6($\downarrow$ 12.2)\\
  16&Yotta &71.6 &-\\
  17&sayazzad &68.3 & -\\
  18&itanhisada &67.7 & -\\
  19&NEUer &66.9 &45.0($\downarrow$ 21.9)\\
  20&YaA@JUST &66.1 & -\\
  21&NLP-IIS@UT &64.4 & -\\
  22&CCLAB &48.1 &31.8($\downarrow$ 16.3)\\
  23&K-FUT &47.6 & -\\
  24&owlmx &44.8 &31.0($\downarrow$13.8 )\\
  25&UIT-ISE-NLP &42.0 &27.3($\downarrow$ 14.7)\\
  26&UoR &39.1 &34.2($\downarrow$ 4.9)\\
  27&Noor &19.9 & -\\
  28&BaoShanCollege &17.6 & -\\
  \hline
\end{tabular}
\vspace*{-2mm}
\caption{Official results of Subtask 2 and Subtask 3. \textit{Acc} is the accuracy (\%) of the models trained on the Subtask 2 training data and tested on the Subtask 2 testset. \textit{Acc. Cross} is the accuracy(\%) of models trained on the \mbox{Subtask 1} training data and tested on the \mbox{Subtask 2} testset.}\label{tab:subtask_2_results}
\end{table}

Many teams used external knowledge resources to further improve model performance. WordNet~\cite{wordnet1998} was widely used to provide candidate word definitions. \texttt{ECNU-ICA-1}~\cite{ecnu} also used ConceptNet5~\cite{speer2016conceptnet} and Graph Neural Network in their systems. To alleviate the noise induced by incorporating structured knowledge through unimportant edges, they propose a noise reduction strategy.
\texttt{owlmx} used the MRC Psycholinguistic Database to obtain a measurement of \textit{imperceptibility} abstractness.

Different pre-processing techniques were proposed in multiple systems. \texttt{ZJUKLAB}~\cite{zju} used a sliding window to limit input length in training. \texttt{PINGAN Omini-Sinitic}~\cite{pingan} used the cycle noisy label detection algorithm to make models more robust. 

Much interesting analysis regarding the failure cases and data distribution was discussed in several system description papers.
\texttt{XRJL}~\cite{hkust} found that for a few questions, common sense knowledge was further needed to help find the answer.
They also pointed out that there were still a few questions in which multiple candidate choices may serve as appropriate answers.

\subsection{Subtask 2: ReCAM-Nonspecificity}
In Subtask 2, we received 28 submissions. Table~\ref{tab:subtask_2_results} shows the official leaderboard. 
The best result in Subtask 2 was achieved by team \texttt{PINGAN omini-Sinitic}~\cite{pingan} with an accuracy of 0.953, using a model similar to the team's model in Subtask 1. The second-placed team \texttt{SRC-B-roc}~\cite{sansung} also adopted the same model it used in Subtask 1 with a data augmentation method based on the hypernym hierarchy in WordNet.

In general, the participating teams in \mbox{Subtask 2} used pre-trained language models and neural networks similar to those they used in \mbox{Subtask 1}. The main differences lie in how the participants performed data augmentation and leveraged external knowledge. For example, in addition to \texttt{SRC-B-roc}~\cite{sansung}, the \texttt{IRG} team~\cite{lrg} also performed data augmentation using hypernyms from WordNet.

\subsection{Subtask 3: Cross-task Performance}
\label{sec:subtask3}
In this section, we explore models' performance across the two types of definitions of abstractness. Specifically, in this subtask, participants train their models on the training set of one subtask and test on the testset of the other subtask. We received 29 submissions in total from the participants.

\paragraph{Cross-task performance: Subtask 2-to-1 testing.}
We asked participants to test their models trained on the Subtask 2 training data on the Subtask 1 test data. The results are shown in the last column of Table~\ref{tab:subtask_1_results}.

The results we received show that the performance of all systems drops substantially. For some systems ranking among top 10, the accuracy can decrease by 5 points (\texttt{IIE-NLP-Eyas}~\cite{iie} and \texttt{XRJL}~\cite{hkust}), or even more (14 points for \texttt{nxc}). Some systems show good generalization ability in this Subtask 2-to-1 scenario; the performance of  \texttt{PINGAN-Omini-Sinitic}~\cite{pingan} is only 1.3 point less, which may be due to the the data augmentation and task adaptive training used in the model. 


\paragraph{Cross-task Performance: Subtask 1-to-2 Testing.} 
Participants are asked to test their Subtask 1 systems on the Subtask 2 testset. Details of the results can be seen in the last column of Table~\ref{tab:subtask_2_results}. All systems' performances drop. For example, among the top-10 systems, the accuracy decreases by 5 points (\texttt{IIE-NLP-Eyas}~\cite{iie}) or 7 points (\texttt{tt123}).  


However, \texttt{ECNU-ICA-1}~\cite{ecnu} shows a very good generalization ability in Subtask 1-to-2 testing.
\texttt{PINGAN-Omini-Sinitic}~\cite{pingan}, \texttt{SRC-B-roc}~\cite{sansung} and \texttt{XRJL}~\cite{hkust}'s systems are rather consistent in this Subtask 1-to-2 cross testing.
Some algorithms they used may explain the models' good generalization ability.  
\texttt{ECNU-ICA-1}'s algorithm of using knowledge-enhanced Graph Attention Network can provide external knowledge to the model. The Wrong Answer Ensemble algorithm~\cite{kim2020learning} used in \texttt{PINGAN-Omini-Sinitic}~\cite{pingan} is a relatively simple but an effective way of improving model performance and generalization ability. Also, the Stochastic Weight Averaging algorithm across multiple checkpoints is effective for better generalization. 
\texttt{XRJL}~\cite{hkust} retrieves the definitions of candidate answers from WordNet and feeds them to the model as extra inputs. We also think data augmentation methods contribute to the generalization ability.


\section{Related Work} 
\label{sec:related_work}
There have been tasks being proposed to evaluate machines' ability on reading comprehension, which either require models to find an entity or text span from the source document as the answer \cite{karl2015teach, hill2015cbtest, onishi2016did, rajpurkar2016squad, trischler2016newsqa}, or further generate an answer \cite{nguyen2016ms, he2017dureader, kovcisky2018narrativeqa}. The cloze-style MRC tasks \cite{karl2015teach,onishi2016did, hill2015cbtest} are most similar to ours, in which the missing words in the cloze questions are entities appearing in source documents. Unlike previous work, ReCAM questions specifically focus on abstract words unseen in the corresponding source documents.

In general, multi-choice questions have been widely used as a tool for language examination to test both humans and machines. In this paper, we follow the multiple-choice framework for our proposed ReCAM task to evaluate computers' ability in comprehending abstract concepts, in which computers are asked to predict the missing abstract words in human-written summaries.

\section{Summary}
This shared task aims to study the ability of machines in representing and understanding abstract concepts, based on two definitions of abstractness, the \textit{imperceptibility} and \textit{nonspecificity}, in a specific machine reading comprehension setup.
We provide three subtasks to evaluate models' ability in comprehending the two types of abstract meaning as well as their generalizability.
In Subtask 1, the top system achieves an accuracy of 0.951, and in Subtask 2, an accuracy of 0.953, suggesting the current systems perform well in the specific setup of our share task. In Subtask 3, we found that  in general the models' performances dropped in both Subtask 2-to-1 and Subtask 1-to-2 testing. However, some models generalize well, benefiting from technologies such as data augmentation and task adaptive training. We hope the shared task can help shed some light on modelling abstract concepts and help design more challenging tasks in the future. 




\bibliographystyle{acl_natbib}
\bibliography{acl2021}

\begin{thebibliography}{45}
\expandafter\ifx\csname natexlab\endcsname\relax\def\natexlab#1{#1}\fi

\bibitem[{Banerjee and Pedersen(2002)}]{satan2002alesk}
Satanjeev Banerjee and Ted Pedersen. 2002.
\newblock \href {https://doi.org/10.1007/3-540-45715-1\_11} {An adapted lesk
  algorithm for word sense disambiguation using wordnet}.
\newblock In \emph{Computational Linguistics and Intelligent Text Processing,
  Third International Conference, CICLing 2002, Mexico City, Mexico, February
  17-23, 2002, Proceedings}, volume 2276 of \emph{Lecture Notes in Computer
  Science}, pages 136--145. Springer.

\bibitem[{Bird and Loper(2004)}]{bird2004nltk}
Steven Bird and Edward Loper. 2004.
\newblock {NLTK}: the natural language toolkit.
\newblock In \emph{Proceedings of the ACL 2004 on Interactive poster and
  demonstration sessions}, page~31. Association for Computational Linguistics.

\bibitem[{Changizi(2008)}]{mark2008hierarchies}
Mark~A. Changizi. 2008.
\newblock \href {https://doi.org/10.1016/j.cogsys.2008.02.001} {Economically
  organized hierarchies in wordnet and the oxford english dictionary}.
\newblock \emph{Cogn. Syst. Res.}, 9(3):214--228.

\bibitem[{Cho et~al.()Cho, Gulcehre, Bahdanau, Schwenk, and
  Bengio}]{cholearning}
Kyunghyun Cho, Bart van Merri{\"e}nboer~Caglar Gulcehre, Dzmitry Bahdanau,
  Fethi Bougares~Holger Schwenk, and Yoshua Bengio.
\newblock Learning phrase representations using rnn encoder--decoder for
  statistical machine translation.

\bibitem[{Clark et~al.(2020)Clark, Luong, Le, and Manning}]{electra}
Kevin Clark, Minh{-}Thang Luong, Quoc~V. Le, and Christopher~D. Manning. 2020.
\newblock \href {https://openreview.net/forum?id=r1xMH1BtvB} {{ELECTRA:}
  pre-training text encoders as discriminators rather than generators}.
\newblock In \emph{8th International Conference on Learning Representations,
  {ICLR} 2020, Addis Ababa, Ethiopia, April 26-30, 2020}. OpenReview.net.

\bibitem[{Coltheart(1981)}]{coltheart1981mrc}
Max Coltheart. 1981.
\newblock The {MRC} psycholinguistic database.
\newblock \emph{The Quarterly Journal of Experimental Psychology},
  33(4):497--505.

\bibitem[{Craswell(2009)}]{craswell2009mean}
Nick Craswell. 2009.
\newblock Mean reciprocal rank.
\newblock In \emph{Encyclopedia of Database Systems}, pages 1703--1703.
  Springer.

\bibitem[{Devlin et~al.(2019)Devlin, Chang, Lee, and Toutanova}]{bert}
Jacob Devlin, Ming{-}Wei Chang, Kenton Lee, and Kristina Toutanova. 2019.
\newblock \href {https://doi.org/10.18653/v1/n19-1423} {{BERT:} pre-training of
  deep bidirectional transformers for language understanding}.
\newblock In \emph{Proceedings of the 2019 Conference of the North American
  Chapter of the Association for Computational Linguistics: Human Language
  Technologies, {NAACL-HLT} 2019, Minneapolis, MN, USA, June 2-7, 2019, Volume
  1 (Long and Short Papers)}, pages 4171--4186. Association for Computational
  Linguistics.

\bibitem[{Dhingra et~al.(2017)Dhingra, Liu, Yang, Cohen, and
  Salakhutdinov}]{dhingra2016gated}
Bhuwan Dhingra, Hanxiao Liu, Zhilin Yang, William~W. Cohen, and Ruslan
  Salakhutdinov. 2017.
\newblock \href {https://doi.org/10.18653/v1/P17-1168} {Gated-attention readers
  for text comprehension}.
\newblock In \emph{Proceedings of the 55th Annual Meeting of the Association
  for Computational Linguistics, {ACL} 2017, Vancouver, Canada, July 30 -
  August 4, Volume 1: Long Papers}, pages 1832--1846. Association for
  Computational Linguistics.

\bibitem[{Fellbaum(1998)}]{wordnet1998}
Christiane Fellbaum. 1998.
\newblock \emph{WordNet: An Electronic Lexical Database}.
\newblock Bradford Books.

\bibitem[{He et~al.(2020)He, Liu, Gao, and Chen}]{deberta}
Pengcheng He, Xiaodong Liu, Jianfeng Gao, and Weizhu Chen. 2020.
\newblock \href {http://arxiv.org/abs/2006.03654} {{DeBERTa}: Decoding-enhanced
  {BERT} with disentangled attention}.
\newblock \emph{CoRR}, abs/2006.03654.

\bibitem[{He et~al.(2018)He, Liu, Liu, Lyu, Zhao, Xiao, Liu, Wang, Wu, She,
  Liu, Wu, and Wang}]{he2017dureader}
Wei He, Kai Liu, Jing Liu, Yajuan Lyu, Shiqi Zhao, Xinyan Xiao, Yuan Liu,
  Yizhong Wang, Hua Wu, Qiaoqiao She, Xuan Liu, Tian Wu, and Haifeng Wang.
  2018.
\newblock \href {https://doi.org/10.18653/v1/W18-2605} {Dureader: a chinese
  machine reading comprehension dataset from real-world applications}.
\newblock In \emph{Proceedings of the Workshop on Machine Reading for Question
  Answering@ACL 2018, Melbourne, Australia, July 19, 2018}, pages 37--46.
  Association for Computational Linguistics.

\bibitem[{Hermann et~al.(2015)Hermann, Kocisk{\'{y}}, Grefenstette, Espeholt,
  Kay, Suleyman, and Blunsom}]{karl2015teach}
Karl~Moritz Hermann, Tom{\'{a}}s Kocisk{\'{y}}, Edward Grefenstette, Lasse
  Espeholt, Will Kay, Mustafa Suleyman, and Phil Blunsom. 2015.
\newblock \href
  {https://proceedings.neurips.cc/paper/2015/hash/afdec7005cc9f14302cd0474fd0f3c96-Abstract.html}
  {Teaching machines to read and comprehend}.
\newblock In \emph{Advances in Neural Information Processing Systems 28: Annual
  Conference on Neural Information Processing Systems 2015, December 7-12,
  2015, Montreal, Quebec, Canada}, pages 1693--1701.

\bibitem[{Hill et~al.(2016)Hill, Bordes, Chopra, and Weston}]{hill2015cbtest}
Felix Hill, Antoine Bordes, Sumit Chopra, and Jason Weston. 2016.
\newblock \href {http://arxiv.org/abs/1511.02301} {The goldilocks principle:
  Reading children's books with explicit memory representations}.
\newblock In \emph{4th International Conference on Learning Representations,
  {ICLR} 2016, San Juan, Puerto Rico, May 2-4, 2016, Conference Track
  Proceedings}.

\bibitem[{Izmailov et~al.(2018)Izmailov, Podoprikhin, Garipov, Vetrov, and
  Wilson}]{izmailov2018averaging}
Pavel Izmailov, Dmitrii Podoprikhin, Timur Garipov, Dmitry~P. Vetrov, and
  Andrew~Gordon Wilson. 2018.
\newblock \href {http://auai.org/uai2018/proceedings/papers/313.pdf} {Averaging
  weights leads to wider optima and better generalization}.
\newblock In \emph{Proceedings of the Thirty-Fourth Conference on Uncertainty
  in Artificial Intelligence, {UAI} 2018, Monterey, California, USA, August
  6-10, 2018}, pages 876--885. {AUAI} Press.

\bibitem[{Jiang et~al.(2021)Jiang, Shou, Wang, Wu, and Lin}]{hkust}
Yuxin Jiang, Ziyi Shou, Qijun Wang, Hao Wu, and Fangzhen Lin. 2021.
\newblock {XRJL-HKUST} at {SemEval-2021 Task 4}: Wordnet-enhanced dual
  multi-head co-attention for reading comprehension of abstract meaning.
\newblock In \emph{Proceedings of the 15th International Workshop on Semantic
  Evaluation (SemEval-2021)}.

\bibitem[{Kim and Fung(2020)}]{kim2020learning}
Hyeondey Kim and Pascale Fung. 2020.
\newblock \href {https://aaai.org/ojs/index.php/AAAI/article/view/7194}
  {Learning to classify the wrong answers for multiple choice question
  answering (student abstract)}.
\newblock In \emph{The Thirty-Fourth {AAAI} Conference on Artificial
  Intelligence, {AAAI} 2020, The Thirty-Second Innovative Applications of
  Artificial Intelligence Conference, {IAAI} 2020, The Tenth {AAAI} Symposium
  on Educational Advances in Artificial Intelligence, {EAAI} 2020, New York,
  NY, USA, February 7-12, 2020}, pages 13843--13844. {AAAI} Press.

\bibitem[{Kingma and Ba(2015)}]{kingma2014adam}
Diederik~P. Kingma and Jimmy Ba. 2015.
\newblock \href {http://arxiv.org/abs/1412.6980} {Adam: {A} method for
  stochastic optimization}.
\newblock In \emph{3rd International Conference on Learning Representations,
  {ICLR} 2015, San Diego, CA, USA, May 7-9, 2015, Conference Track
  Proceedings}.

\bibitem[{Ko{\v{c}}isk{\`y} et~al.(2018)Ko{\v{c}}isk{\`y}, Schwarz, Blunsom,
  Dyer, Hermann, Melis, and Grefenstette}]{kovcisky2018narrativeqa}
Tom{\'a}{\v{s}} Ko{\v{c}}isk{\`y}, Jonathan Schwarz, Phil Blunsom, Chris Dyer,
  Karl~Moritz Hermann, G{\'a}abor Melis, and Edward Grefenstette. 2018.
\newblock The narrativeqa reading comprehension challenge.
\newblock \emph{Transactions of the Association of Computational Linguistics},
  6:317--328.

\bibitem[{Lan et~al.(2020)Lan, Chen, Goodman, Gimpel, Sharma, and
  Soricut}]{albert}
Zhenzhong Lan, Mingda Chen, Sebastian Goodman, Kevin Gimpel, Piyush Sharma, and
  Radu Soricut. 2020.
\newblock \href {https://openreview.net/forum?id=H1eA7AEtvS} {{ALBERT:} {A}
  lite {BERT} for self-supervised learning of language representations}.
\newblock In \emph{8th International Conference on Learning Representations,
  {ICLR} 2020, Addis Ababa, Ethiopia, April 26-30, 2020}. OpenReview.net.

\bibitem[{Lesk(1986)}]{lesk1986automatic}
Michael Lesk. 1986.
\newblock Automatic sense disambiguation using machine readable dictionaries:
  how to tell a pine cone from an ice cream cone.
\newblock In \emph{Proceedings of the 5th annual international conference on
  Systems documentation}, pages 24--26. ACM.

\bibitem[{Liu et~al.(2021)Liu, Wang, Zhao, Chen, Feng, Lin, and He}]{ecnu}
Pingsheng Liu, Linlin Wang, Qian Zhao, Hao Chen, Yuxi Feng, Xin Lin, and Liang
  He. 2021.
\newblock {ECNU\_ICA\_1 SemEval-2021 Task 4}: Leveraging knowledge-enhanced
  graph attention networks for reading comprehension of abstract meaning.
\newblock In \emph{Proceedings of the 15th International Workshop on Semantic
  Evaluation (SemEval-2021)}.

\bibitem[{Liu et~al.(2019)Liu, Ott, Goyal, Du, Joshi, Chen, Levy, Lewis,
  Zettlemoyer, and Stoyanov}]{roberta}
Yinhan Liu, Myle Ott, Naman Goyal, Jingfei Du, Mandar Joshi, Danqi Chen, Omer
  Levy, Mike Lewis, Luke Zettlemoyer, and Veselin Stoyanov. 2019.
\newblock \href {http://arxiv.org/abs/1907.11692} {{RoBERTa}: {A} robustly
  optimized {BERT} pretraining approach}.
\newblock \emph{CoRR}, abs/1907.11692.

\bibitem[{Miller(1998)}]{miller1998wordnet}
George Miller. 1998.
\newblock \emph{WordNet: An electronic lexical database}.
\newblock MIT press.

\bibitem[{Moro et~al.(2014)Moro, Raganato, and Navigli}]{moro2014entity}
Andrea Moro, Alessandro Raganato, and Roberto Navigli. 2014.
\newblock Entity linking meets word sense disambiguation: a unified approach.
\newblock \emph{Transactions of the Association for Computational Linguistics},
  2:231--244.

\bibitem[{Narayan et~al.(2018)Narayan, Cohen, and Lapata}]{shashi2018xsum}
Shashi Narayan, Shay~B. Cohen, and Mirella Lapata. 2018.
\newblock \href {https://doi.org/10.18653/v1/d18-1206} {Don't give me the
  details, just the summary! topic-aware convolutional neural networks for
  extreme summarization}.
\newblock In \emph{Proceedings of the 2018 Conference on Empirical Methods in
  Natural Language Processing, Brussels, Belgium, October 31 - November 4,
  2018}, pages 1797--1807. Association for Computational Linguistics.

\bibitem[{Nguyen et~al.(2016)Nguyen, Rosenberg, Song, Gao, Tiwary, Majumder,
  and Deng}]{nguyen2016ms}
Tri Nguyen, Mir Rosenberg, Xia Song, Jianfeng Gao, Saurabh Tiwary, Rangan
  Majumder, and Li~Deng. 2016.
\newblock \href {http://ceur-ws.org/Vol-1773/CoCoNIPS\_2016\_paper9.pdf} {{MS}
  {MARCO:} {A} human generated machine reading comprehension dataset}.
\newblock In \emph{Proceedings of the Workshop on Cognitive Computation:
  Integrating neural and symbolic approaches 2016 co-located with the 30th
  Annual Conference on Neural Information Processing Systems {(NIPS} 2016),
  Barcelona, Spain, December 9, 2016}, volume 1773 of \emph{{CEUR} Workshop
  Proceedings}. CEUR-WS.org.

\bibitem[{Onishi et~al.(2016)Onishi, Wang, Bansal, Gimpel, and
  McAllester}]{onishi2016did}
Takeshi Onishi, Hai Wang, Mohit Bansal, Kevin Gimpel, and David McAllester.
  2016.
\newblock Who did what: A large-scale person-centered cloze dataset.
\newblock In \emph{Proceedings of the 2016 Conference on Empirical Methods in
  Natural Language Processing}, pages 2230--2235.

\bibitem[{Pennington et~al.(2014)Pennington, Socher, and
  Manning}]{pennington2014glove}
Jeffrey Pennington, Richard Socher, and Christopher Manning. 2014.
\newblock Glove: Global vectors for word representation.
\newblock In \emph{Proceedings of the 2014 conference on empirical methods in
  natural language processing (EMNLP)}, pages 1532--1543.

\bibitem[{Pruksachatkun et~al.(2020)Pruksachatkun, Phang, Liu, Htut, Zhang,
  Pang, Vania, Kann, and Bowman}]{pruksachatkun2020intermediate}
Yada Pruksachatkun, Jason Phang, Haokun Liu, Phu~Mon Htut, Xiaoyi Zhang,
  Richard~Yuanzhe Pang, Clara Vania, Katharina Kann, and Samuel~R. Bowman.
  2020.
\newblock \href {https://doi.org/10.18653/v1/2020.acl-main.467}
  {Intermediate-task transfer learning with pretrained language models: When
  and why does it work?}
\newblock In \emph{Proceedings of the 58th Annual Meeting of the Association
  for Computational Linguistics, {ACL} 2020, Online, July 5-10, 2020}, pages
  5231--5247. Association for Computational Linguistics.

\bibitem[{Qi et~al.(2020)Qi, Zhang, Zhang, Bolton, and Manning}]{qi2020stanza}
Peng Qi, Yuhao Zhang, Yuhui Zhang, Jason Bolton, and Christopher~D. Manning.
  2020.
\newblock \href {http://arxiv.org/abs/2003.07082} {Stanza: A {Python} natural
  language processing toolkit for many human languages}.

\bibitem[{Raffel et~al.(2020)Raffel, Shazeer, Roberts, Lee, Narang, Matena,
  Zhou, Li, and Liu}]{t5}
Colin Raffel, Noam Shazeer, Adam Roberts, Katherine Lee, Sharan Narang, Michael
  Matena, Yanqi Zhou, Wei Li, and Peter~J. Liu. 2020.
\newblock \href {http://jmlr.org/papers/v21/20-074.html} {Exploring the limits
  of transfer learning with a unified text-to-text transformer}.
\newblock \emph{J. Mach. Learn. Res.}, 21:140:1--140:67.

\bibitem[{Rajpurkar et~al.(2016)Rajpurkar, Zhang, Lopyrev, and
  Liang}]{rajpurkar2016squad}
Pranav Rajpurkar, Jian Zhang, Konstantin Lopyrev, and Percy Liang. 2016.
\newblock \href {https://doi.org/10.18653/v1/d16-1264} {Squad: 100, 000+
  questions for machine comprehension of text}.
\newblock In \emph{Proceedings of the 2016 Conference on Empirical Methods in
  Natural Language Processing, {EMNLP} 2016, Austin, Texas, USA, November 1-4,
  2016}, pages 2383--2392. The Association for Computational Linguistics.

\bibitem[{Sanh et~al.(2019)Sanh, Debut, Chaumond, and Wolf}]{distilbert}
Victor Sanh, Lysandre Debut, Julien Chaumond, and Thomas Wolf. 2019.
\newblock {DistilBERT}, a distilled version of {BERT}: smaller, faster, cheaper
  and lighter.
\newblock \emph{arXiv preprint arXiv:1910.01108}.

\bibitem[{Sharma et~al.(2021)Sharma, Pandey, Chhablani, Bhartia, and
  Dash}]{lrg}
Abheesht Sharma, Harshit Pandey, Gunjan Chhablani, Yash Bhartia, and Tirtharaj
  Dash. 2021.
\newblock {LRG} at {SemEval-2021} {Task 4}: Improving reading comprehension
  with abstract words using augmentation, linguistic features and voting.
\newblock In \emph{Proceedings of the 15th International Workshop on Semantic
  Evaluation (SemEval-2021)}.

\bibitem[{Speer et~al.(2016)Speer, Chin, and Havasi}]{speer2016conceptnet}
Robert Speer, Joshua Chin, and Catherine Havasi. 2016.
\newblock \href {http://arxiv.org/abs/1612.03975} {Conceptnet 5.5: An open
  multilingual graph of general knowledge}.
\newblock In \emph{AAAI Conference on Artificial Intelligence}.

\bibitem[{Spreen and Schulz(1966)}]{spreen1966abstract}
Otfried Spreen and Rudolph~W. Schulz. 1966.
\newblock \href {https://doi.org/https://doi.org/10.1016/S0022-5371(66)80061-0}
  {Parameters of abstraction, meaningfulness, and pronunciability for 329
  nouns}.
\newblock \emph{Journal of Verbal Learning and Verbal Behavior}, 5(5):459 --
  468.

\bibitem[{Theijssen et~al.(2011)Theijssen, Halteren, Boves, and
  Oostdijk}]{theijssen2011difficulty}
DL~Theijssen, H~van Halteren, LWJ Boves, and NHJ Oostdijk. 2011.
\newblock On the difficulty of making concreteness concrete.

\bibitem[{Trischler et~al.(2017)Trischler, Wang, Yuan, Harris, Sordoni,
  Bachman, and Suleman}]{trischler2016newsqa}
Adam Trischler, Tong Wang, Xingdi Yuan, Justin Harris, Alessandro Sordoni,
  Philip Bachman, and Kaheer Suleman. 2017.
\newblock \href {https://doi.org/10.18653/v1/w17-2623} {Newsqa: {A} machine
  comprehension dataset}.
\newblock In \emph{Proceedings of the 2nd Workshop on Representation Learning
  for NLP, Rep4NLP@ACL 2017, Vancouver, Canada, August 3, 2017}, pages
  191--200. Association for Computational Linguistics.

\bibitem[{Turney et~al.(2011)Turney, Neuman, Assaf, and
  Cohen}]{turney2011literal}
Peter~D Turney, Yair Neuman, Dan Assaf, and Yohai Cohen. 2011.
\newblock Literal and metaphorical sense identification through concrete and
  abstract context.
\newblock In \emph{Proceedings of the Conference on Empirical Methods in
  Natural Language Processing}, pages 680--690. Association for Computational
  Linguistics.

\bibitem[{Wang et~al.(2021)Wang, Wang, Zhu, Zeng, Hao, Wang, and Xiao}]{pingan}
Ye~Wang, Yanmeng Wang, Haijun Zhu, Bo~Zeng, Zhenghong Hao, Shaojun Wang, and
  Jing Xiao. 2021.
\newblock {PINGAN Omini-Sinitic} at {SemEval-2021} {Task 4}: Reading
  comprehension of abstract meaning.
\newblock In \emph{Proceedings of the 15th International Workshop on Semantic
  Evaluation (SemEval-2021)}.

\bibitem[{Xie et~al.(2021{\natexlab{a}})Xie, Chen, Chen, Wang, Zhang, Deng, and
  Chen}]{zju}
Xin Xie, Xiangnan Chen, Xiang Chen, Yong Wang, Ningyu Zhang, Shumin Deng, and
  Huajun Chen. 2021{\natexlab{a}}.
\newblock {ZJUKLAB} at {S}em{E}val-2021 {Task 4}: Negative augmentation with
  language model for reading comprehension of abstract meaning.
\newblock In \emph{Proceedings of the 15th International Workshop on Semantic
  Evaluation (SemEval-2021)}.

\bibitem[{Xie et~al.(2021{\natexlab{b}})Xie, Xing, Peng, and Hu}]{iie}
Yuqiang Xie, Luxi Xing, Wei Peng, and Yue Hu. 2021{\natexlab{b}}.
\newblock {IIE-NLP-Eyas} at {SemEval-2021} {Task 4}: Enhancing plm for recam
  with special tokens, re-ranking, siamese encoders and back translation.
\newblock In \emph{Proceedings of the 15th International Workshop on Semantic
  Evaluation (SemEval-2021)}.

\bibitem[{Yang et~al.(2019)Yang, Dai, Yang, Carbonell, Salakhutdinov, and
  Le}]{xlnet}
Zhilin Yang, Zihang Dai, Yiming Yang, Jaime~G. Carbonell, Ruslan Salakhutdinov,
  and Quoc~V. Le. 2019.
\newblock \href
  {https://proceedings.neurips.cc/paper/2019/hash/dc6a7e655d7e5840e66733e9ee67cc69-Abstract.html}
  {Xlnet: Generalized autoregressive pretraining for language understanding}.
\newblock In \emph{Advances in Neural Information Processing Systems 32: Annual
  Conference on Neural Information Processing Systems 2019, NeurIPS 2019,
  December 8-14, 2019, Vancouver, BC, Canada}, pages 5754--5764.

\bibitem[{Zhang et~al.(2021)Zhang, Zhuang, and Su}]{sansung}
Jing Zhang, Yimeng Zhuang, and Yinpei Su. 2021.
\newblock {TA-MAMC} at {SemEval-2021} {Task4}: Task-adaptive pretraining and
  multi-head attention for abstract meaning reading comprehension.
\newblock In \emph{Proceedings of the 15th International Workshop on Semantic
  Evaluation (SemEval-2021)}.

\end{thebibliography}

\newpage
\appendix

\section{Annotation Script}
\label{annotation_script}
\includegraphics[scale=0.70]{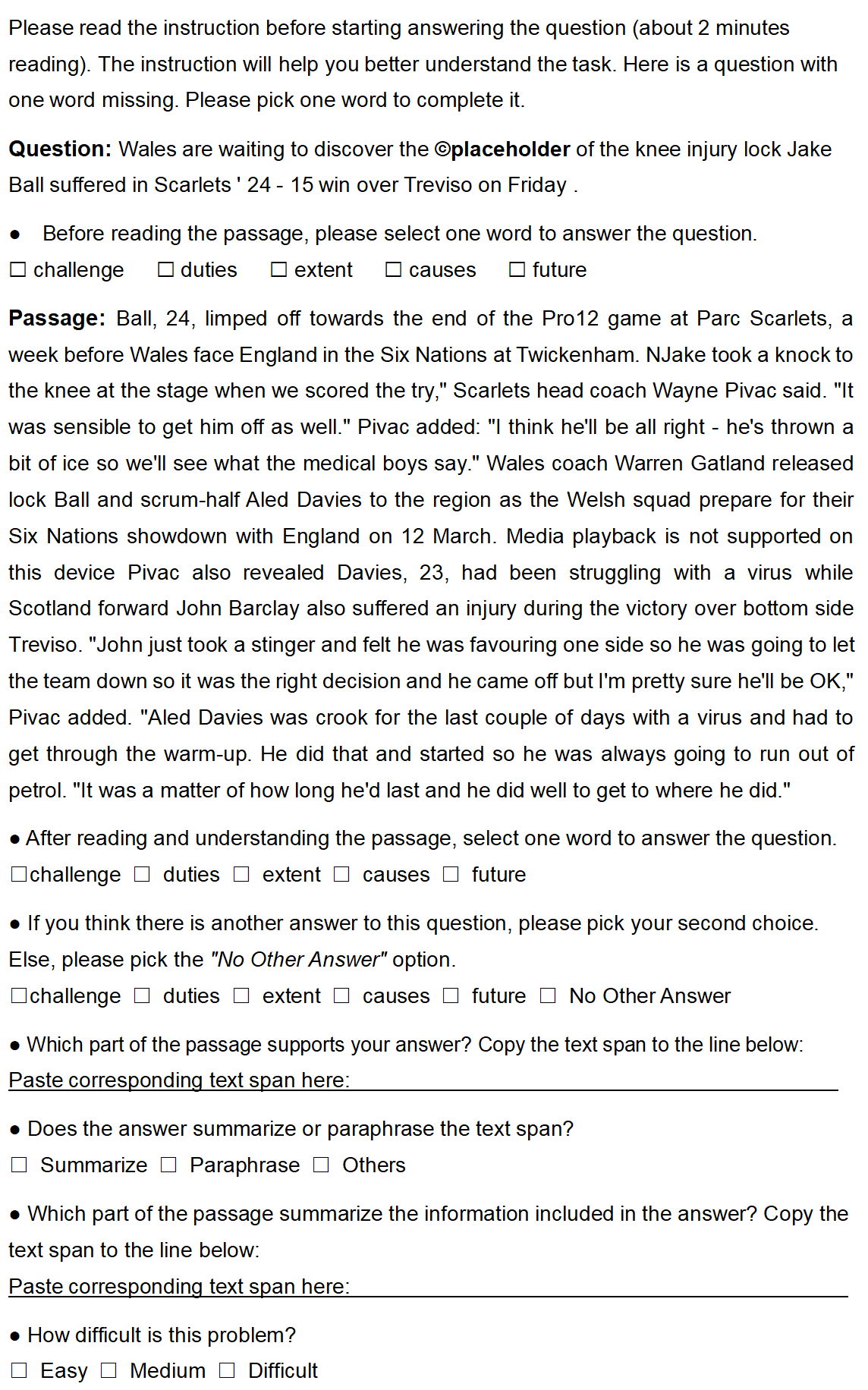}
\clearpage

\section{Gated-Attention Reader}
\label{gated_attention_reader}
The Gated-Attention (GA) Reader \cite{dhingra2016gated}, the state-of-art model on CNN/Daily Mail reading comprehension dataset \cite{karl2015teach}, is adapted here in our experiments. The GA Reader uses a multi-layer iterated architecture with a gated attention mechanism, which is based on multiplicative interactions between the query embedding and the intermediate states of a recurrent neural network document reader, to derive better query-aware passage representation. To apply GA Reader to our ARC task,
we input the news passage $p$ as the document and the processed summary $s$ as the query to GA Reader. 

Specifically, for an input passage $p=[p_1, p_2, ..., p_{l_p}]$ with $l_p$ words and its corresponding summary $s=[s_1, s_2, ..., s_{l_s}]$ with $l_s$ words, we first derive their corresponding word embedding sequence $\mathbf{P}=[\mathbf{p}_1, \mathbf{p}_2, ..., \mathbf{p}_{l_p}]$ and $\mathbf{S}=[\mathbf{s}_1, \mathbf{s}_2, ..., \mathbf{s}_{l_s}]$ respectively. Then the GA Reader accepts the $\mathbf{P}$ and $\mathbf{S}$ as inputs and return the hidden states $\mathbf{H}_p=[\mathbf{h}_1^p, \mathbf{h}_2^p, ..., \mathbf{h}_{l_p}^p]$ and $\mathbf{H}_s=[\mathbf{h}_1^s, \mathbf{h}_2^s, ..., \mathbf{h}_{l_s}^s]$ as the sequential representation for passage $p$ and summary $s$ respectively. As for the final prediction process, we do not adopt the operations in \newcite{dhingra2016gated} because in ARC the answer words are unseen in the corresponding passage, however, GA Reader in \newcite{dhingra2016gated} tries to select a entity word in the passage as the final prediction since their target answer word appears in the passage. So we redesign the part of prediction.

First, the corresponding representation of ``\emph{@placeholder}'' in $\mathbf{H}_s$, denoted as $\mathbf{h}_{q}^s$ ($q$ is the position index of \emph{@placeholder} in summary $s$), is used as the final vector representation for summary $s$. For the final vector representation $\mathbf{p}$ for passage $p$, a bilinear attention between $\mathbf{h}_q^s$ and $\mathbf{H}_p$ is used for its derivation:
\begin{align}
e_i &= {\mathbf{h}_q^s}^T \mathbf{W}^{att} \mathbf{h}_i^p, \forall i \in [1, ..., l_p]\\
 \mathbf{p} &= \sum_{i=1}^{l_p} \frac{\exp{e_{i}}}{\sum_{j=1}^{l_p} \exp{e_{j}}} \mathbf{h}_i^p,
\end{align}
We set a token embedding $\mathbf{a}_t^e$ for each candidate abstractive word $a_t$ ($t\in[1, ..., n_c]$, $n_c$ is the size of candidate set). We first concatenate the $\mathbf{h}_p^s$ and $\mathbf{p}$, then use the bilinear product and softmax to predict the probability distribution over all $n_c$ candidate abstractive words.
\begin{align}
r_t &= [\mathbf{h}_q^s; \mathbf{p}]^T \mathbf{W}_p \mathbf{a}_t^e,  \forall t \in [1, ..., n_c],\\
{o}_t&= {softmax}_t(r_t), \forall t \in [1, ..., n_c]
\end{align}
in which ${o}_t$ represents the probability of predicting the candidate abstractive word 
$a_t$ as the final answer.

\section{Attentive Model}
\label{attentive_model}

\begin{figure*}[!t]
  \centering
  \includegraphics[width=6.2in]{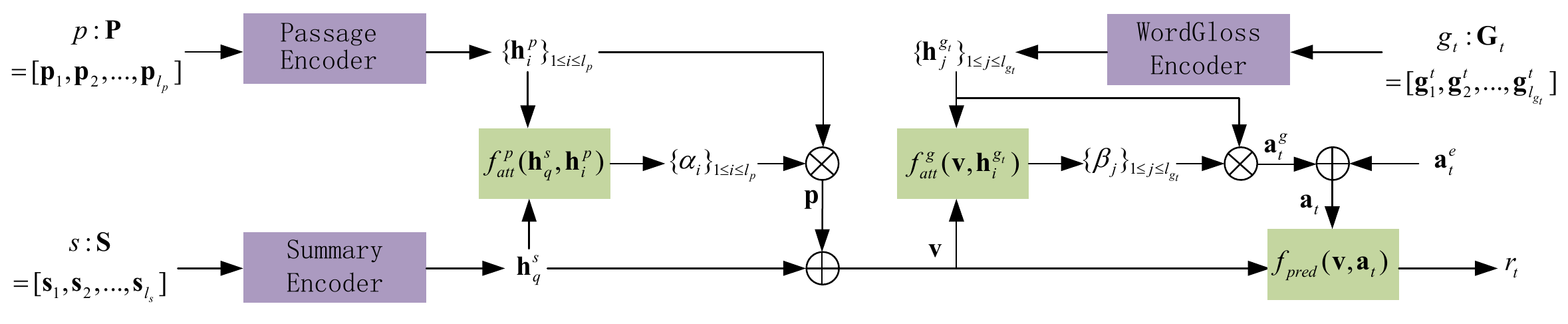}
  \caption{The model architecture of the attentive model with word gloss (AMWG) implemented in this paper. $\bigoplus$ denotes the concatenation of input vectors. All the encoders are 1-layer bi-directional GRU-RNNs, $\bigotimes$ denotes the weighted sum of vectors.}
  \label{fig:models}
\end{figure*}

The word gloss, which defines a word sense meaning, has been mainly used in word sense disambiguation (WSD) task and its variants \cite{lesk1986automatic, moro2014entity}. 
Since the goal of ARC is to predict a word that can summarize corresponding information from the source passage, which is an abstracting process, it may be helpful when the gloss, i.e., interpretation of candidate abstractive words, are provided.


We design an attentive model with word gloss (AMWG) as Figure \ref{fig:models} shows. Specifically, all the \emph{encoders} are 1-layer bi-directional recurrent neural networks (RNNs) with Gated Recurrent Units (GRU) \cite{cholearning}. For an input news passage $p=[p_1, p_2, ..., p_{l_p}]$ with $l_p$ words, we can derive its hidden states $\mathbf{H}_p=[\mathbf{h}_1^p, \mathbf{h}_2^p, ..., \mathbf{h}_{l_p}^p]$ by sending its word embedding sequence $\mathbf{P}=[\mathbf{p}_1, \mathbf{p}_2, ..., \mathbf{p}_{l_p}]$ to the \emph{Passage Encoder}. Similarly, we can derive hidden states  $\mathbf{H}_s=[\mathbf{h}_1^s, \mathbf{h}_2^s, ..., \mathbf{h}_{l_s}^s]$ for summary $s$ by inputting its word embedding sequence $\mathbf{S}=[\mathbf{s}_1, \mathbf{s}_2, ..., \mathbf{s}_{l_s}]$ into the \emph{Summary Encoder} and hidden states $\mathbf{H}_{g_t}=[\mathbf{h}_1^{g_t}, \mathbf{h}_2^{g_t}, ..., \mathbf{h}_{l_{g_t}}^{g_t}]$ for gloss $g_t$ of the candidate word $a_t$ by sending its word embedding sequence $\mathbf{G}_t=[\mathbf{g}_1^t, \mathbf{g}_2^t, ..., \mathbf{g}_{l_{g_t}}^t]$ to the \emph{WordGloss Encoder}.

Similar to Section \ref{gated_attention_reader}, the corresponding representation of ``\emph{@placeholder}'', i.e., $\mathbf{h}_{q}^s$, is used as the final vector representation for summary $s$. And an bilinear attention $f_{att}^p( \bullet )$ is applied to $\mathbf{h}_{q}^s$ and $\mathbf{H}_p$ as follows:
\begin{align}
e_i &= {\mathbf{h}_q^s}^T \mathbf{W}_{att}^p \mathbf{h}_i^p, \forall i \in [1, ..., l_p]\\
 \alpha_i &=  \frac{\exp{e_{i}}}{\sum_{j=1}^{l_p} \exp{e_{j}}}, \forall i \in [1, ..., l_p]
\end{align}
Then $\mathbf{p}$ is derived as the vector representation for passage $p$ by the weighed sum of $\mathbf{H}_p$, which is further concatenated with the $\mathbf{h}_{q}^s$ to form the final summarization vector $\mathbf{v}$:
\begin{align}
\mathbf{p} &= \sum_{i=1}^{l_p} \alpha_i \mathbf{h}_i^p,\\
\mathbf{v} &= concat(\mathbf{p}, \mathbf{h}_{q}^s),
\end{align}
Another attention $f_{att}^{g}(\bullet)$ is applied to $\mathbf{v}$ and $\mathbf{H}_{g_t}$, 
\begin{align}
e_j &= {\tanh(\mathbf{W}_{att}^g \mathbf{v} + \mathbf{b})}^T \mathbf{h}_j^{g_t}, \forall j \in [1, ..., l_{g_t}]\\
\beta_j &= \frac{\exp{e_{j}}}{\sum_{i=1}^{l_{g_t}} \exp{e_{i}}}, \forall j \in [1, ..., l_{g_t}],
\end{align}
The following weighted sum of $\mathbf{H}_{g_t}$, i.e, $\mathbf{a}_t^g$, is derive as the final vector representation for the gloss of candidate word $a_t$:
\begin{align}
\mathbf{a}_t^g &= \sum_{j=1}^{l_{g_t}} \beta_j \mathbf{h}_j^{g_t}
\end{align}
We also set a token embedding $\mathbf{a}_t^e$ for each candidate word $a_t$ ($t\in[1, ..., n_c]$, $n_c$ is the size of candidate set), which is further concatenated with $\mathbf{a}_t^g$ to build the final representation $\mathbf{a}_t$ for candidate word $a_t$. For the final prediction, we input the summarization vector $\mathbf{v}$ and candidate representation vector $\mathbf{a}_t$ to $f_{pred}(\bullet)$ and apply the softmax to derive the probability  distribution over all $n_c$ candidate abstractive words,
\begin{align}
\mathbf{a}_t &= concat(\mathbf{a}_t^g, \mathbf{a}_t^e),\\
r_t &= \mathbf{v}^T \mathbf{W}_{pred} \mathbf{a}_t,  \forall t \in [1, ..., n_c],\\
{o}_t&= {softmax}_t(r_t), \forall t \in [1, ..., n_c]
\end{align}
in which ${o}_t$ gives the probability of predicting the candidate word $a_t$ as the final answer.
The word gloss, which defines a word sense meaning, has been mainly used in word sense disambiguation (WSD) task and its variants \cite{lesk1986automatic, moro2014entity}. 
Since the goal of ARC is to predict a word that can summarize corresponding information from the source passage, which is an abstracting process, it may be helpful when the gloss, i.e., interpretation of candidate abstractive words, are provided.


We design an attentive model with word gloss (AMWG) as Figure \ref{fig:models} shows. Specifically, all the \emph{encoders} are 1-layer bi-directional recurrent neural networks (RNNs) with Gated Recurrent Units (GRU) \cite{cholearning}. For an input news passage $p=[p_1, p_2, ..., p_{l_p}]$ with $l_p$ words, we can derive its hidden states $\mathbf{H}_p=[\mathbf{h}_1^p, \mathbf{h}_2^p, ..., \mathbf{h}_{l_p}^p]$ by sending its word embedding sequence $\mathbf{P}=[\mathbf{p}_1, \mathbf{p}_2, ..., \mathbf{p}_{l_p}]$ to the \emph{Passage Encoder}. Similarly, we can derive hidden states  $\mathbf{H}_s=[\mathbf{h}_1^s, \mathbf{h}_2^s, ..., \mathbf{h}_{l_s}^s]$ for summary $s$ by inputting its word embedding sequence $\mathbf{S}=[\mathbf{s}_1, \mathbf{s}_2, ..., \mathbf{s}_{l_s}]$ into the \emph{Summary Encoder} and hidden states $\mathbf{H}_{g_t}=[\mathbf{h}_1^{g_t}, \mathbf{h}_2^{g_t}, ..., \mathbf{h}_{l_{g_t}}^{g_t}]$ for gloss $g_t$ of the candidate word $a_t$ by sending its word embedding sequence $\mathbf{G}_t=[\mathbf{g}_1^t, \mathbf{g}_2^t, ..., \mathbf{g}_{l_{g_t}}^t]$ to the \emph{WordGloss Encoder}.

Similar to Section \ref{gated_attention_reader}, the corresponding representation of ``\emph{@placeholder}'', i.e., $\mathbf{h}_{q}^s$, is used as the final vector representation for summary $s$. And an bilinear attention $f_{att}^p( \bullet )$ is applied to $\mathbf{h}_{q}^s$ and $\mathbf{H}_p$ as follows:
\begin{align}
e_i &= {\mathbf{h}_q^s}^T \mathbf{W}_{att}^p \mathbf{h}_i^p, \forall i \in [1, ..., l_p]\\
 \alpha_i &=  \frac{\exp{e_{i}}}{\sum_{j=1}^{l_p} \exp{e_{j}}}, \forall i \in [1, ..., l_p]
\end{align}
Then $\mathbf{p}$ is derived as the vector representation for passage $p$ by the weighed sum of $\mathbf{H}_p$, which is further concatenated with the $\mathbf{h}_{q}^s$ to form the final summarization vector $\mathbf{v}$:
\begin{align}
\mathbf{p} &= \sum_{i=1}^{l_p} \alpha_i \mathbf{h}_i^p,\\
\mathbf{v} &= concat(\mathbf{p}, \mathbf{h}_{q}^s),
\end{align}
Another attention $f_{att}^{g}(\bullet)$ is applied to $\mathbf{v}$ and $\mathbf{H}_{g_t}$, 
\begin{align}
e_j &= {\tanh(\mathbf{W}_{att}^g \mathbf{v} + \mathbf{b})}^T \mathbf{h}_j^{g_t}, \forall j \in [1, ..., l_{g_t}]\\
\beta_j &= \frac{\exp{e_{j}}}{\sum_{i=1}^{l_{g_t}} \exp{e_{i}}}, \forall j \in [1, ..., l_{g_t}],
\end{align}
The following weighted sum of $\mathbf{H}_{g_t}$, i.e, $\mathbf{a}_t^g$, is derive as the final vector representation for the gloss of candidate word $a_t$:
\begin{align}
\mathbf{a}_t^g &= \sum_{j=1}^{l_{g_t}} \beta_j \mathbf{h}_j^{g_t}
\end{align}
We also set a token embedding $\mathbf{a}_t^e$ for each candidate word $a_t$ ($t\in[1, ..., n_c]$, $n_c$ is the size of candidate set), which is further concatenated with $\mathbf{a}_t^g$ to build the final representation $\mathbf{a}_t$ for candidate word $a_t$. For the final prediction, we input the summarization vector $\mathbf{v}$ and candidate representation vector $\mathbf{a}_t$ to $f_{pred}(\bullet)$ and apply the softmax to derive the probability  distribution over all $n_c$ candidate abstractive words,
\begin{align}
\mathbf{a}_t &= concat(\mathbf{a}_t^g, \mathbf{a}_t^e),\\
r_t &= \mathbf{v}^T \mathbf{W}_{pred} \mathbf{a}_t,  \forall t \in [1, ..., n_c],\\
{o}_t&= {softmax}_t(r_t), \forall t \in [1, ..., n_c]
\end{align}
in which ${o}_t$ gives the probability of predicting the candidate word $a_t$ as the final answer.

\section{Training Details}
\label{training_details}
We train all models using the non-negative log-likelihood as the objective function. The gloss of candidate words are derived from \emph{WordNet} using the NLTK tools \cite{bird2004nltk}. Specifically, we first lemmatize the candidate word and use the lemmatized word as the query word for the searching in \emph{WordNet}. To cope with the semantic ambiguity of words, we just concatenate the gloss of the first sense in each retrieved POS for the query word with corresponding POS tag as the deliminator.

Models in our experiments are trained with the following hyperparameter settings: All word embeddings and token embeddings $\mathbf{a}_t^e$ have $300$ dimensions and are initialized with Glove \cite{pennington2014glove}. The passage $p$ and summary $s$ share one set of word embeddings, which are fixed during training. The glosses $\{g_t\}$ for candidate words $\{a_t\}$ keep its own word embeddings. The hidden state vectors of all bi-directional GRU-RNNs in all models have $150$ dimensions. The number of attention hops in GA Reader is set to $3$. The batch size is set to $32$. The method of Adam \cite{kingma2014adam} is adopted for optimization with initial learning rate $1e-03$. A dropout with rate $0.3$ is applied to the input layers for all GRU-RNN encoders and the final summarization vector $\mathbf{v}$.

\section{Annotation Selection}
\label{annotation_selection}
To ensure most of our annotation is valid, we select annotations satisfying the following criteria: a) the average accuracy is higher than 40\%; b) both text spans should not be empty; c) if the difficulty level is rated as easy, then this data sample should be answered correctly. 

\end{document}